\documentclass{article} 
\usepackage{iclr2015,times}
\usepackage{hyperref}
\usepackage{url}
\usepackage[pdftex]{graphicx}
\usepackage{subfig}

\title{Using Neural Networks for Click Prediction of Sponsored Search}

\author{
Afroze I.~Baqapuri \thanks{Most of the work was done while the author was an intern at Yandex.} \\
School of Computer and Communications Sciences\\
\`{E}cole Polytechnique F\`{e}d\`{e}rale de Lausanne\\
Lausanne, Switzerland \\
\texttt{afroze.baqapuri@alumni.epfl.ch} \\
\And
Ilya Trofimov \\
Machine Learning Group \\
Yandex \\
Moscow, Russia \\
\texttt{trofim@yandex-team.ru} \\
}

\begin{document}

\maketitle

\begin{abstract}
Sponsored search is a multi-billion dollar industry and makes up a major source of revenue for search engines (SE). click-through-rate (CTR) estimation plays a crucial role for ads selection, and greatly affects the SE revenue, advertiser traffic and user experience. We propose a novel architecture for solving CTR prediction problem by combining artificial neural networks (ANN) with decision trees. First we compare ANN with respect to other popular machine learning models being used for this task. Then we go on to combine ANN with MatrixNet (proprietary implementation of boosted trees) and evaluate the performance of the system as a whole. The results show that our approach provides significant improvement over existing models.
\end{abstract}

\section{Introduction}

\textit{Sponsored search} results are the small textual advertisements displayed on the search engine. Before displaying the ads are ranked by the SE according to two aspects: ad's relevance to the context so that more users will click on the ad, and some measure of the expected payment received from the advertiser. We limit ourselves to the most common cost-per-click model (CPC): advertiser is charged the bid price each time a user clicks the ad. Therefore, ad's CTR multiplied with the advertiser's bid is recognized as the estimated revenue. That is why CTR estimation plays an integral role in sponsored search and SE revenue.

In this study we use ANNs to model the click through rate of sponsored search. We propose a two-stage click-prediction system, which incorporates ANN into the existing framework of decision trees currently used at Yandex. Up to our knowledge, only \citet{rnn} have   used   ANNs   in   the   domain   of   click prediction. They use recurrent neural networks to model dependency on user’s sequential behavior in sponsored search. Compared to them our approach uses feed forward neural networks and models ID-based input features relating to the user, keyword, query, and advertisement. It should be noted that our paper presents a first step towards using ANNs at a large scale inside Yandex, and there still needs to be much work done in this domain.

Modern search engines use machine learning approaches to predict the CTR. Popular models include logistic regression (LR) \citep{Richardson2007,McMahan2013,yahoo} and boosted decision trees \citep{Dembczynski,trofimov}. ANNs have advantage over LR because they are able to capture non-linear relationship between the input features and their "deeper" architecture has inherently greater modeling strength. On the other hand decision trees - albeit popular in this domain - face additional challenges with  with high-dimensional and sparse ID-based features.  We consider ANNs to be promising models because they already  show state-of-the-art  results  in  various  other  domains  including computer vision \citep{ImageNet}, natural language processing \citep{Collobert2011} and speech recognition \citep{Speech}.

The rest of this paper is organized as follows: In section 2 we briefly describe the existing search advertisement framework at Yandex. In Section 3 we discuss the experimental setting including the input features and the proposed model. In section 4 we present and discuss our results, and in section 5 we conclude our paper and propose some future work.

\section{Sponsored Search Framework in Yandex}

Mechanism of sponsored search is based on keyword auction: advertiser bids on a selected set of keywords. When a user types a query, the SE matches it with all keywords and selects appropriate ads to display. A simplified algorithm for selecting ads is described as follows. First all ads which match the user's query are selected and sorted in descending order according to expected revenue. Then the leading ads - at most three - are picked and sorted according to their bids.

CTR is essential to this process as it is used in calculating the expected revenue. At Yandex \textit{MatrixNet} is the machine learning algorithm used for estimating the CTR. MatrixNet is a proprietary implementation of boosted decision trees and it is successfully applied to numerous numerous classification and regression problems at the company. For sponsored search we use MatrixNet with real-valued features derived from click data logs. The input features describe statistics relating to the following groups: user, context, query, keyword, advertisement and advertiser. For more details about these input features refer to the paper by \citet{trofimov}.

\section{Experimental Setting}

\subsection{Input Features}

We use click-through logs of Yandex search engine as our data set comprising of about 6.6 million training examples. The data was collected during one week between July 1 and July 7, 2014. For this study we focus only on ID-based features belonging to 14 namespaces, which can be divided into following categories:

\begin{itemize}
\item \textbf{User}: User ID, region ID.\\
\item \textbf{Ad}: Ad ID, campaign ID, domain ID, words of ad title, words ad body, ad position, ad keywords.\\
\item \textbf{Query}: Words of user query
\end{itemize}

These features are encoded in 1-of-c encoding which results in very high dimensionality and extremely sparse features space. Since it would be infeasible to directly input the data directly into the neural network, we first attempt to reduce its dimensionality. This is done in two steps:

\begin{enumerate}
\item \textbf{Infrequent feature removal}: If a feature occurs less than a certain threshold, we simply discard it. We found 10 to be a good threshold, which effectively reduces number of unique features from 8.5 million to 1.1 million.\\
\item \textbf{Hashing trick}: We use a hash function to map the remaining features into a lower dimensional space, resulting in a compressed representation of the original feature space. There are bound to be some collisions, but  \citet{yahoo} empirically showed that it is not a major concern. We fix the hashing space dimensionality to 100,000.
\end{enumerate}

It should be noted that since LR models have much fewer parameters, we do not perform these simplification steps on data input to LR. Following this we randomly divide the data into 70\% training (4.6 M), 20 \% validation (1.3 M) and 10\% testing (660 K) sets. An important point to note is that data has highly imbalanced classes (roughly 90\% of the ad impressions resulting in no clicks). We cannot treat it as a standard classification problem because of high unfair bias towards negative class. Instead, we treat it as a stochastic process where the output of the model gives the probability of the ad being clicked.


\subsection{Proposed Model}

We design the click prediction system as a two-stage process. In the first stage ANN models the sparse high-dimensional ID-based features. The second stage is MatrixNet which models the real-valued features. For each ad impression the ANN outputs a real-valued probability of that ad being clicked. This value serves as one of the input features of the MatrixNet. The other inputs are the ones being currently used as input features at Yandex and are described by \citet{trofimov}. The output of MatrixNet gives the final CTR of the ad impression which can be used to estimate the expected revenue. This two stage system is used for two reasons. Firstly, it provides a way of efficiently combining the two kinds of features (real-valued and ID-based) for CTR prediction. Secondly, since Yandex already uses MatrixNet for the task, this is the easiest way to incorporate the ANN  model into the existing framework without any major overhaul. We also train an LR model as a baseline for comparison with ANN.

Negative log likelihood (NLL) is used as the error criterion for training the models, while NLL and and area under precision / recall curve (auPRC) are used for evaluation on the test set.

\subsection{Model Optimization}

We train the LR with L-BFGS \citep{Nocedal1980}, which is a state of the art second-order optimizer. BFGS is a popular choice for training logistic regression and is used by \citet{yahoo} and \citet{Richardson2007}. A review of different optimization methods is presented by \citet{Minka2003} which shows BFGS to be fast and perform well in practice on logistic regression. We used Vowpal Wabbit (VW) \citep{VowpalWabbit} to build and train these linear models. The hyper parameters to optimize are the number of input bits (VW has a built in hashing function for reducing features) and l2 regularization parameter. As stated before we do not deliberately attempt to cut the LR input dimensionality and we do a grid search to find the combination which gives the best results on test set, and use that as our baseline comparison. It took on average 0.36 hours to fully train an LR model so it was reasonably easy to perform this grid search. We also attempted using quadratic features to model non-linearity but - although it took much longer to train the model - the NLL on test set rose by at least 2.4\% (without restricting the input feature space). Hence in all further experiments and results we use simple features.

We train the ANN with stochastic gradient descent (SGD) with mini-batches of size 100. We observed that using a smaller batch size improved results slightly but significantly increased training time, so we decided to fix it to the reasonable size of 100 for all experiments. The hyper parameters which require optimization are learning rate, l2 regularization parameter, number of hidden layers, sizes of hidden layers. We use held-out validation set as a termination criterion for training the network. On average it took us between 55-140 hours to fully train the network. This tremendous increase over LR training time is mainly because number of network parameters (weights) are much larger and it took much more epochs to reach optimization on the validation set. To add on to this, there was no available framework available to us to parallelize training on multiple cores for extremely sparse layers of the ANN. We used torch7 for building and training the ANN models \citep{torch7}.

Initially it was quite hard to optimize ANN model to good performance since we observed a high sensitivity to learning rate and l2 regularization parameter. However inclusion of learning rate decay and dropout seemed to somewhat alleviate this problem. We also discovered that using rectified linear units (relu) as non-linearity function was instrumental to the success of the model. In fact ANN models trained without relu and dropout were performing worse about the same as, or even worse than best LR model. Using \textit{relu} resulted in up to 0.87\% reduction in NLL and 5.7\% increase in auPRC. While using dropout gave significant improvements of up to 0.96\% reduction in NLL and 5.36\% increase in auPRC. We searched several combinations along these hyper parameters and found the best combination as follows: l2 coefficient = 3e-4; learning rate = 0.1; learning rate decay = 2e-4 per million training instances; non-linearity function = rectified linear units. 

It may be worth mentioning that the input data has inherent divisibility, since the sparse features can be grouped into 14 namespaces. We tried to make use of this and experimented on a network with local connectivity. To be precise, we separated the input features in their respective namespaces and then applied the dimensionality reduction techniques to each namespace separately so that that there is no conflict between different namespaces. The number of resulting features in each namespace was in proportion to the number of original features in that, such that the total features still equalled 100,000. After this we fed the 14 inputs to 14 smaller neural networks separately, and concatenated the outputs of these deeper in the network. The advantage of this approach was that it considerably reduced the number of parameters in the ANN model, while also reduced the possibility of conflict between features of different namespaces in the input layer. However, the experiments performed with it resulted in NLL error increase by up to 0.33\% so we did not consider it in our further experiments.

\section{Results}

\subsection{Individual ANN}

First we evaluate the ANN independently, not considering the MatrixNet stage. The softmax output of ANN can be interpreted as CTR which is used for evaluating the neural network performance. We select the best six ANN architectures using early stopping on validation set. Table \ref{mlp-layer-comparison} shows the changes in NLL and auPRC of these models over the baseline LR performance. It is clear that replacing LR with ANN considerably improves performance, with the best network architecture showing 0.88\% improvement in log likelihood and 5.57\% improvement in auPRC metric.

Experimentation also revealed that using rectified linear units (relu) was instrumental for achieving this performance improvement. Using \textit{relu} resulted in up to 0.87\% improvement in log likelihood and up to 5.7\% improvement in auPRC. Similarly, using dropout functionality in the hidden units of the network enhanced the performance.

\begin{table}[t]
\caption{Results of 6 best ANN models (percentage change over LR results).}
\label{mlp-layer-comparison}
\begin{center}
\begin{tabular}{|c|c|c|}
\hline
\bf{Models}  & \bf{$\Delta$ NLL}        & \bf{$\Delta$ auPRC}     \\ \hline
1 h-layer ANN - 10 hU       &    -0.58\%    & 3.77\%    \\ \hline
1 h-layer ANN - 25 hU        &     -0.48\%   & 4.68\%    \\ \hline
1 h-layer ANN - 50 hU         &    -0.88\%   & 5.57\%    \\ \hline
1 h-layer ANN - 100 hU         &   -0.65\%   & 5.38\%    \\ \hline
2 h-layer ANN - 50 + 50 hU      &  -0.75\%   & 4.98\%    \\ \hline
2 h-layer ANN - 100 + 100 hU     & -0.66\%   & 4.69\%    \\ \hline
\end{tabular}
\end{center}
\end{table}

\subsection{Ensemble of ANNs}

In ensemble methods results of many models are averaged to give performance improvement. The principle is that if different models settle down to very different local optimal solutions then they will give incorrect predictions for different sets of instances. So if we combine and average the predictions of these models before comparing them to the target of the test instance, it may result in better performance as compared to the results of any individual model. We experiment with this method and use the outputs of the best six ANN models by averaging them together.

\begin{figure}[ht]
\begin{center}
\subfloat[]{\includegraphics[width=2.5in]{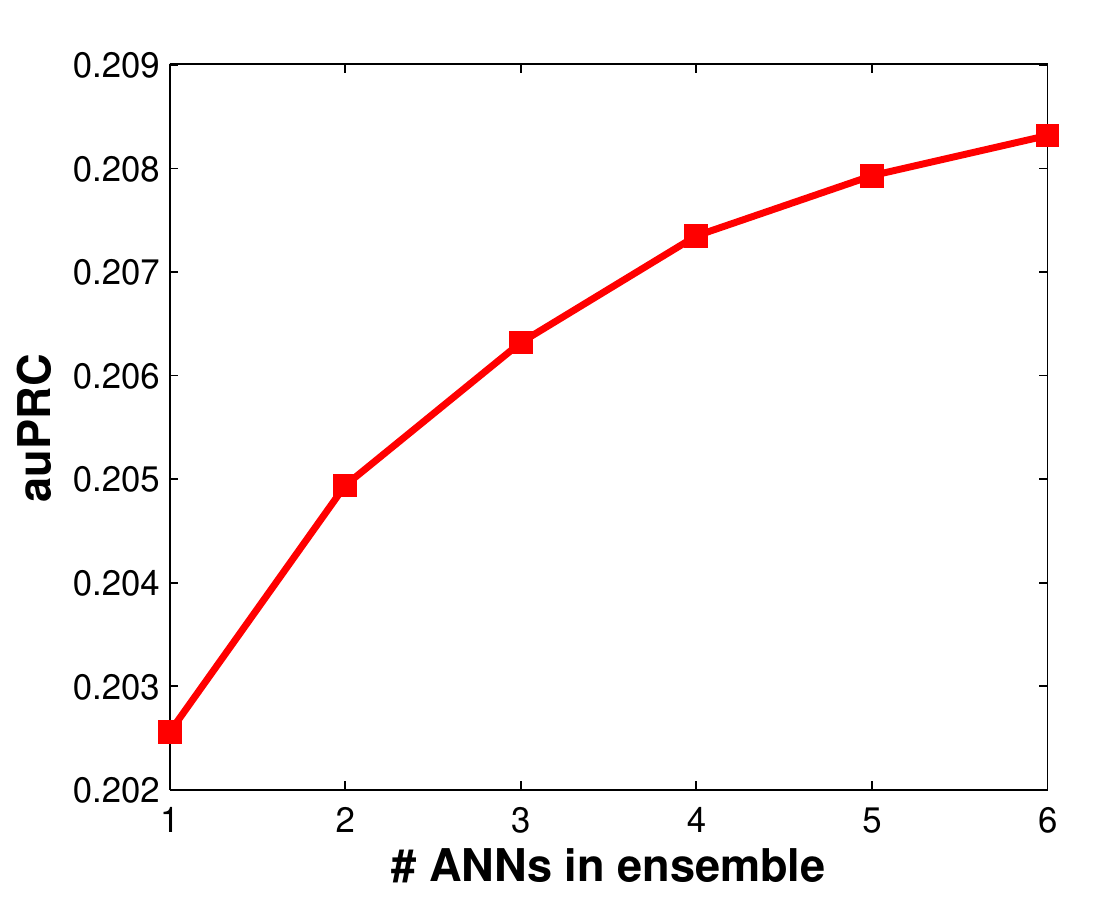}}
\subfloat[]{\includegraphics[width=2.5in]{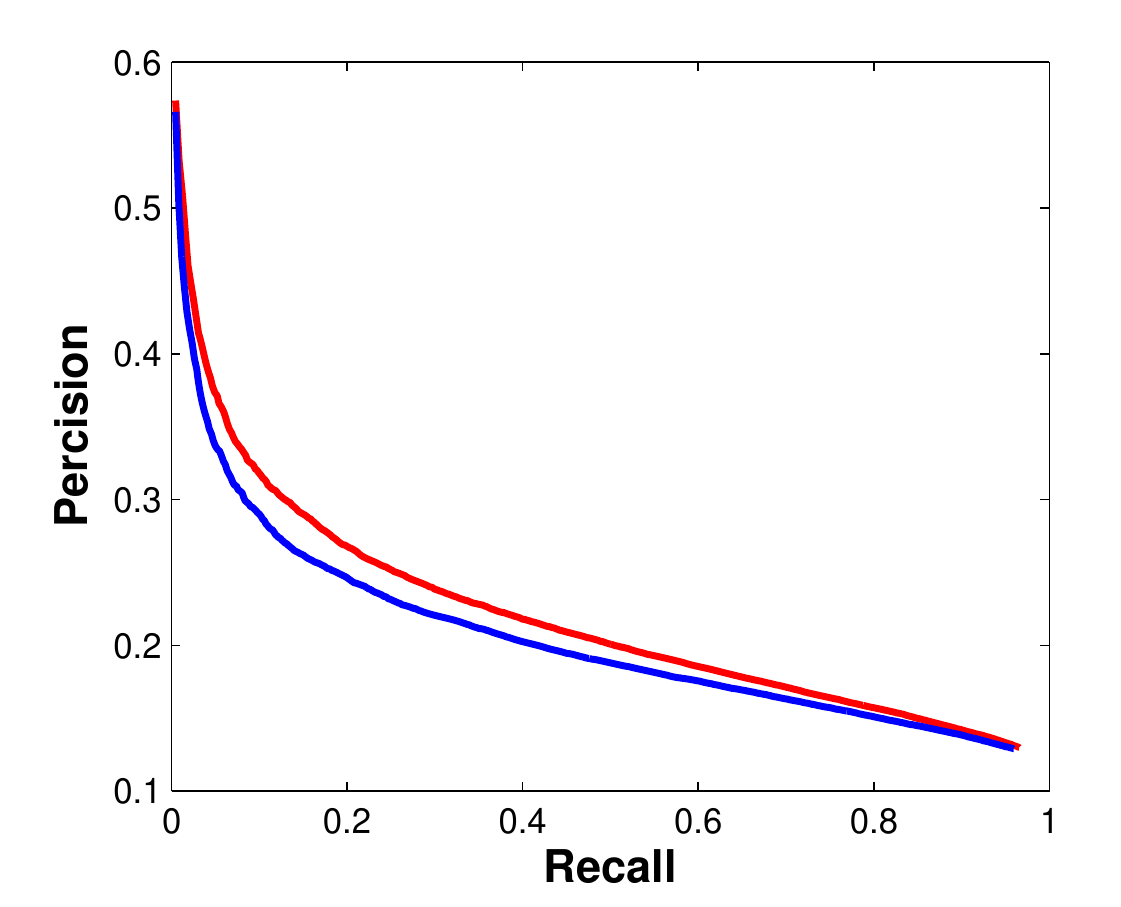}}
\end{center}
\caption{(a) Effect of increasing the number of ANNs in the ensemble on the auPRC. (b) Precision-recall curves for LR (blue) and ensemble (red) show improvement in auPRC by using ensemble.}
\label{ensemble-comparison}
\end{figure}

Figure \ref{ensemble-comparison}(a) shows that performance improves steadily as we keep on increasing the number of ANNs in our ensemble. ANN models are added to the ensemble in the same order as they are listed in Table \ref{mlp-layer-comparison} (moving from top to bottom).  In Table \ref{final-comparison} we compare the performance of the ensemble and the best performing individual ANN with the LR results. We can observe that using ensemble improves the log likelihood by 0.24\% and auPRC by 1.15\% over the individual ANN. Figure \ref{ensemble-comparison}(b) plots the precision-recall curves for the baseline LR and ensemble model. It clearly shows a visible improvement in auPRC for the ensemble.

\begin{table}[t]
\caption{Comparison of best ANN results with baseline LR.}
\label{final-comparison}
\begin{center}
\begin{tabular}{|c|c|c|c|}
\hline
\bf{Model Category}  & \bf{$\Delta$ NLL}    & \bf{auPRC}    & \bf{$\Delta$ auPRC}     \\ \hline
logistic Regression     & 0\%           & 0.19520           & 0\%       \\ \hline
Individual ANN          & -0.88\%       & 0.20608           & 5.57\%       \\ \hline
Ensemble of 6 ANNs      & -1.12\%       & 0.20832           & 6.72\%       \\ \hline
\end{tabular}
\end{center}
\end{table}

\subsection{Size of Training Set}

\begin{figure}[ht]
\begin{center}
\includegraphics[width=2.5in]{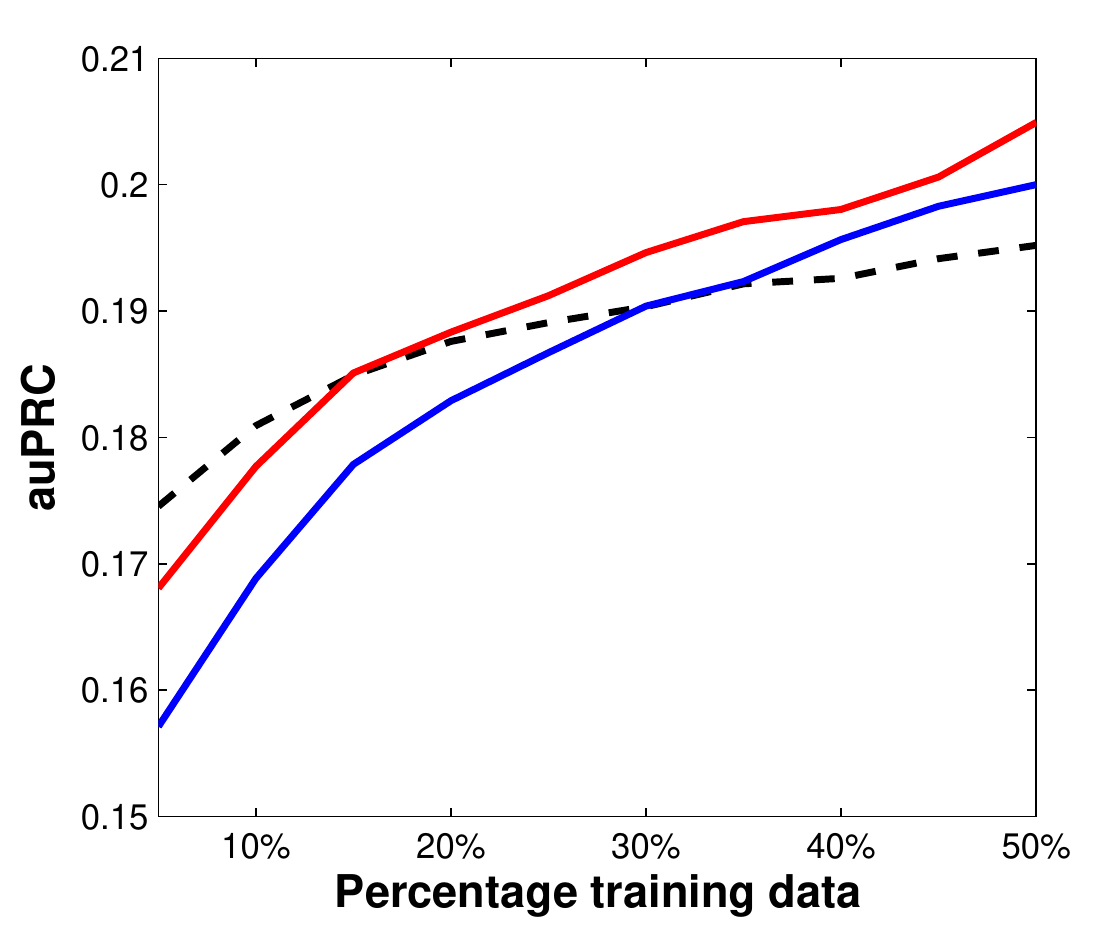}
\end{center}
\caption{Effect of the change in size of training data on the LR (dashed), 1 hidden-layer ANN (blue) and 2 hidden-layers ANN (red).}
\label{sizeratio-auc}
\end{figure}

We wanted to determine the effect of training data size on the performance. For this purpose we train the ANN models on subsets of our training data (not changing the test and validation sets). Figure \ref{sizeratio-auc} shows the result of this experiment, where different models are evaluated and compared using auPRC metric. As expected, increasing the training size improves performance of all models. However, we notice that when the training data is small LR outperforms ANN. But as the training data increases, ANN catches up to LR, and in fact the performance gap keeps increasing steadily in favour of ANN. This means that true potential of ANN is exploited when using large data sets. From the plots we can infer that if we use even more training data, ANN would perform even better than LR as compared to current results.   


\subsection{MatrixNet Results}

Now we will evaluate the complete click-prediction system at Yandex as a whole. As stated before in Section 3, this consists of ANN followed by MatrixNet. For comparison we will also evaluate the existing framework: MatrixNet alone with baseline features without ANN output. The comparison results in Table \ref{matrixnet-comparison} show that using ANN output as additional input to MatrixNet causes up to 0.22\% improvement in  log likelihood, and 0.48\% improvement in auPRC. All the results are calculated on a separate test set. From our experience we know that results with a difference in log likelihood of 0.1\% is small but important, and should not be neglected \citep{trofimov}. Due to this observation we consider that using neural networks in the click prediction system at Yandex results in significant improvement.

\begin{table}[t]
\caption{Results of different models' outputs as additional input to MatrixNet (percentage change over original MatrixNet features).}
\label{matrixnet-comparison}
\begin{center}
\begin{tabular}{|c|c|c|c|}
\hline
\bf{Model}  & \bf{$\Delta$ NLL}                 & \bf{auPRC}      & \bf{$\Delta$ auPRC}     \\ \hline
MatrixNet with baseline features    & 0\%       & 0.26073       & 0\%    \\ \hline
MatrixNet + Logistic Regression     & -0.07\%   & 0.26115       & 0.16\% \\ \hline
MatrixNet + Individual ANN          & -0.20\%   & 0.26200       & 0.48\% \\ \hline
MatrixNet + Ensemble of 6 ANNs      & -0.22\%   & 0.26191       & 0.45\% \\ \hline
\end{tabular}
\end{center}
\end{table}

\section{Conclusion and Future Work}

In this paper we proposed using ANNs for modeling ID-based features for CTR prediction task. First we showed that using non-linear models like ANN improves performance over linear model like LR. We went on to show that using an ensemble of ANNs improves performance even more and also using more data further increases the performance gap between . Finally we did a comparison and stated improvements of using ANN in combination with existing click prediction framework MatrixNet.

This paper presents a first step towards using ANN inside Yandex for click prediction. Further research can include testing on real-time data, and see the performance effects on a real-time ad selection system. However, more work would need to be done on improving time efficiency of the ANN system with extremely sparse input (which does not parallelize well with multiple cores). As the results clearly show gap in performance improving with large data size, it would be interesting to see the effect of using much larger training data. Moreover, since many of the ID-based features are in forms of words it may be useful to initialize the neural network as an RBM trained with unsupervised contrastive divergence on a large volume of unlabeled examples. And then fine tune it as a discriminative model with back propagation. It could also prove useful to train multiple networks in parallel and feed all of their outputs individually to MatrixNet, as feature vectors, instead of just a single average.

\subsubsection*{Acknowledgments}

We would like to thank the Machine Learning group and the Computer Vision group at Yandex for helpful discussions.

\bibliography{iclr2015}
\bibliographystyle{iclr2015}

\end{document}